%% file: paper.tex
\documentclass[]{bytedance_seed}

\usepackage[toc,page,header]{appendix}

\input{iclr2026/math_commands}

\usepackage{hyperref}
\usepackage{url}
\usepackage{algorithm}
\usepackage{algorithmic}
\usepackage{enumitem}
\usepackage{fvextra}
\usepackage{multirow}
\usepackage{booktabs}
\usepackage[table]{xcolor}
\usepackage{hyperref}
\usepackage{wrapfig}
\usepackage{graphicx} 
\usepackage{fvextra}
\usepackage{url}

\usepackage{minitoc}
\usepackage{tikz}
\usetikzlibrary{decorations.fractals}
\usepackage{tikz}
\usetikzlibrary{lindenmayersystems}

\pgfdeclarelindenmayersystem{Koch}{
  \rule{F -> F+F--F+F}
}
\pgfdeclarelindenmayersystem{Plant}{
  \rule{X -> F-[[X]+X]+F[+FX]-X}
  \rule{F -> FF}
}
\pgfdeclarelindenmayersystem{Dragon}{
  \rule{X -> X+YF+}
  \rule{Y -> -FX-Y}
}

\newcommand{\decoraten}[2]{%
  \ifnum#1>0
    decorate{ \decoraten{\numexpr#1-1\relax}{#2} }%
  \else
    #2%
  \fi}

\newcommand{\code}[1]{{\ttfamily#1}}

\newcommand{\sgreen}[1]{{\tiny\textcolor{green!50!black}{#1}}}
\newcommand{\sred}[1]{{\tiny\textcolor{red!50!black}{#1}}}

\title{Scaling Long-Horizon 
LLM Agent via Context-Folding}

\author[1,2,*]{Weiwei Sun}
\author[1,3,*]{Miao Lu}
\author[1]{Zhan Ling}
\author[1]{Kang Liu}
\author[1]{Xuesong Yao}
\author[2]{Yiming Yang}
\author[1, \dagger]{Jiecao Chen}

\affiliation[1]{ByteDance Seed}
\affiliation[2]{Carnegie Mellon University}
\affiliation[3]{Stanford University}

\contribution[*]{Work done at ByteDance Seed}
\contribution[\dagger]{Corresponding authors}

\abstract{
Large language model (LLM) agents are fundamentally constrained by context length on long-horizon tasks. 
We introduce Context-Folding, a framework that empowers agents to actively manage their working context. An agent can procedurally branch into a sub-trajectory to handle a subtask and then fold it upon completion, collapsing the intermediate steps while retaining a concise summary of the outcome. To make this behavior learnable, we develop an end-to-end reinforcement learning framework FoldGRPO with specific process rewards to encourage effective task decomposition and context management.  On complex long-horizon tasks (Deep Research and SWE), our folding agent matches or outperforms the ReAct baselines while using an active context 10$\times$ smaller and significantly outperforms models that rely on summarization-based context management.

}

\date{\today}
\correspondence{Weiwei Sun at \email{sunnweiwei@gmail.com}, Jiecao Chen at \email{jiecao.chen@bytedance.com}}
\checkdata[Project Page]{\url{https://context-folding.github.io/}}

\begin{document}

\maketitle


\input{main}

\bibliographystyle{plainnat}
\bibliography{reference}

\clearpage

\beginappendix


\input{appendix}

\end{document}

%% file: iclr2026/math_commands.tex
\usepackage{amsmath,amsfonts,bm,amssymb}
\usepackage{amsthm}
\usepackage{mathtools}

\mathtoolsset{showonlyrefs}

\newtheoremstyle{mytheoremstyle}    
    {}                              
    {}                              
    {\it}                           
    {}                              
    {\bfseries}                     
    {.}                             
    {.5em}                          
    {}                              
\theoremstyle{mytheoremstyle}

\numberwithin{theorem}{section}

\def\1{\bm{1}}

\DeclareMathAlphabet{\mathsfit}{\encodingdefault}{\sfdefault}{m}{sl}
\SetMathAlphabet{\mathsfit}{bold}{\encodingdefault}{\sfdefault}{bx}{n}



%% file: main.tex
\section{Introduction}
Large language model (LLM) agents have shown remarkable capabilities in tackling complex, long-horizon problems that require extensive interaction with an environment, such as deep research~\citep{OpenAI2025a,Google2025b,Jin2025SearchR1TL,Wei2025BrowseCompAS,Li2025WebThinkerEL} and agentic coding~\citep{Jimenez2023SWEbenchCL,AnthropicClaudeCode2025,Wang2024OpenHandsAO}. 
The length of tasks agents can complete is argued to be \textit{growing exponentially, with a doubling time of about 7 months}~\citep{METR2025a}.

However, scaling LLM agents to even longer horizons is fundamentally constrained by the design of agentic frameworks~\citep{Yao2022ReAct}. 
These frameworks linearly accumulate the entire interaction history (reasoning, tool calls, and observations) into a single, ever-expanding context, which incurs long-context challenges as horizons scale: (1) degraded performance, as LLMs struggle to utilize relevant information in exceedingly long contexts~\citep{Liu2023LostIT,Comanici2025Gemini2P,Leng2024LongCR}; and (2) poor efficiency, stemming from the quadratic scaling of attention mechanisms and the growing overhead of managing the KV-cache~\citep{Katharopoulos2020TransformersAR}.

Existing approaches to scale long-horizon LLM agents largely fall into two classes: 
(1) \textit{Summary-based methods}, which trigger a post-hoc summarization stage when the working context is full~\citep{OpenHands2025,Yu2025MemAgentRL,Qiao2025WebResearcherUU,Wu2025ReSumUL,Zhou2025MEM1LT,lu2025scaling}. While this compresses the context, it can abruptly disrupt the agent’s working context and reasoning flow, which may lead to sub-optimal results. 
(2) \textit{Multi-agent systems}, which distribute tasks across specialized agents to manage context length~\citep{Zhao2024LongAgentSL,Zhang2024ChainOA,Anthropic2025c,Wong2025WideSearchBA}. Yet, these systems typically depend on handcrafted, problem-specific workflows that are difficult to generalize and resist end-to-end optimization.

\begin{figure}
    \centering
    \includegraphics[width=1.0\linewidth]{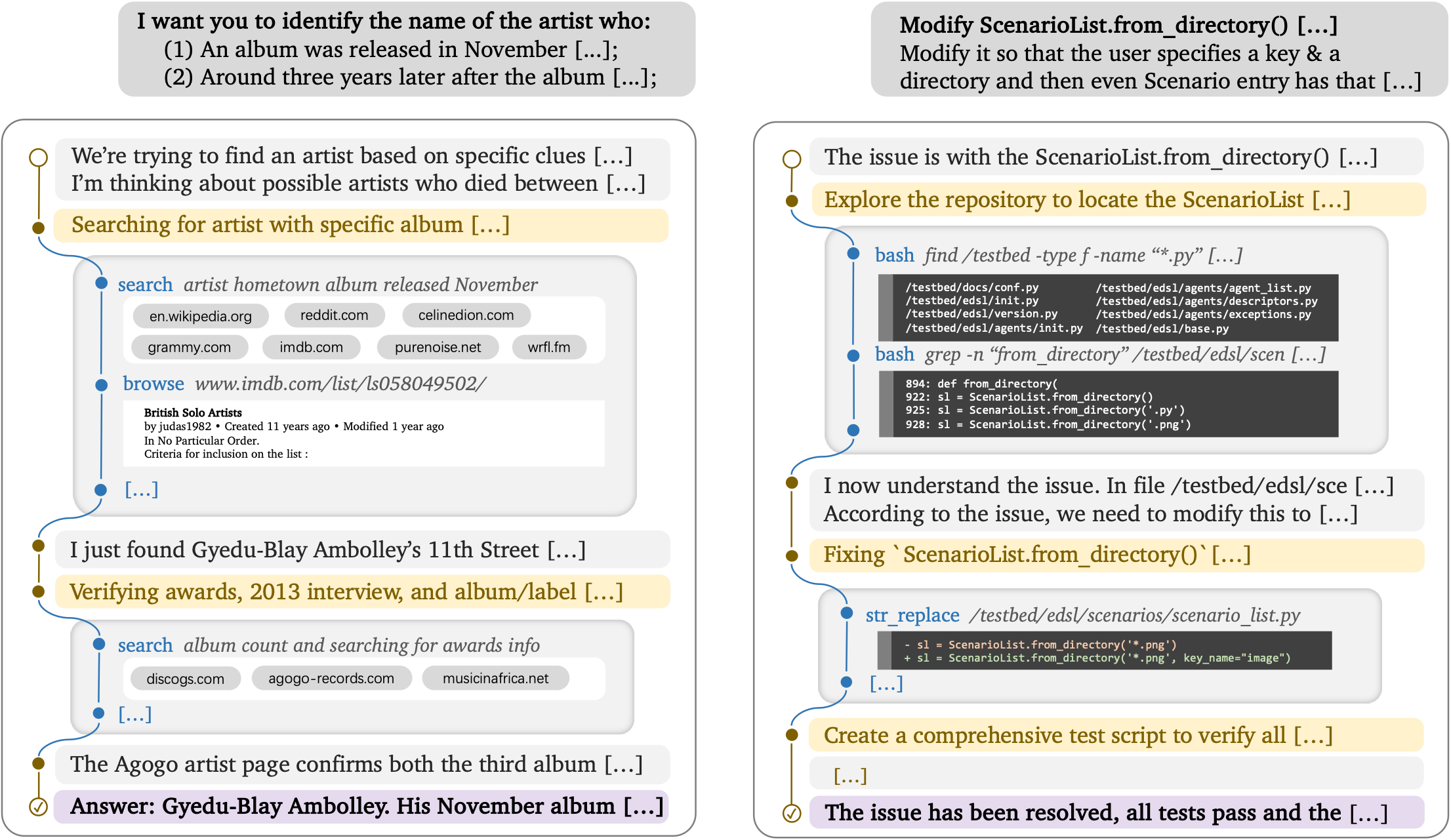}
    \vspace{-0.3cm}
    \caption{Examples of context folding in long-horizon tasks: deep research (left) and agentic coding (right).}\label{fig:example}
\end{figure}

In this paper, we propose \textbf{Context Folding}: an agentic mechanism that allows the model to actively manage its working context. 
Specifically, the agent manages its context using two special actions:
(i) a \texttt{branch} action to create a temporary sub-trajectory for a localized subtask; and
(ii) a \texttt{return} action to summarize the outcome and rejoin the main thread,
after which the intermediate steps within the branch are ``folded''---removed from the context —leaving only a concise summary from the \texttt{return} call. 
Figure~\ref{fig:example} illustrates this process on deep research and agentic coding tasks, where the agent offloads token-intensive actions (e.g., web search or codebase exploration) into branches and preserves only key findings and insights for high-level reasoning.
Compared with existing methods, context folding enables an agentic approach to active context management, where the agent’s short-term context remains undisrupted and long-term context is automatically managed.

Based on the context-folding framework, we propose a novel end-to-end reinforcement learning algorithm for training LLM agents on complex, long-horizon tasks. 
The key innovation is \textbf{FoldGRPO}, which augments the standard GRPO by incorporating (i) dynamic folded LLM contexts and (ii) dense, token-level process rewards that directly guide context folding behavior.
Specifically, our RL algorithm teaches the model how to effectively decompose a problem into localized sub-tasks for branching, guided by an \textit{Unfolded Token Penalty} that discourages token-heavy operations in the main context. Furthermore, it learns to maintain focus within a sub-task via an \textit{Out-of-Scope Penalty}, and to preserve crucial information in its summaries to aid the final objective. 
By mastering these skills, the agent can handle vastly longer interaction histories, allowing our framework to scale the agent's effective horizon and improve overall system efficiency.

We evaluate our approach on two long-horizon benchmarks, BrowseComp-Plus~\citep{Chen2025BrowseCompPlusAM} and SWE-Bench Verified~\citep{Jimenez2023SWEbenchCL}, where our agent achieves strong performance with remarkable efficiency. 
Despite using a compact 32K active token budget managed with maximum of 10 branches, our agent, the \textbf{Folding Agent}, achieves pass@1 scores of 62.0\% and 58.0\% respectively, surpassing baselines that require a massive 327K context window and significantly outperforming methods based on context summarization. 
The effectiveness of our method is rooted in reinforcement learning, which provides absolute improvements of 20.0\% on BrowseComp-Plus and 8.8\% on SWE-Bench. 
Further analysis reveals that our agent learns to invoke more tool calls and generate longer outputs to handle complex problems, and can scale to larger token budgets at inference time to tackle even more challenging tasks. 
Together, these results indicate that learning to \textit{actively manage} context, rather than merely extending or heuristically compressing it, is a principled path toward scalable long-horizon agency.

In summary, our contributions are threefold: 
\begin{enumerate}[label=(\roman*),leftmargin=20pt]
\item We introduce \textbf{Context Folding}, a mechanism that enables agents to actively manage their context and mitigate the challenge of linear history growth.
\item We present \textbf{FoldGRPO}, a reinforcement learning framework with dynamic folded LLM contexts and dense process rewards that trains agents to effectively acquire the capability of context folding.
\item We demonstrate promising performance on long-horizon benchmarks, highlighting our approach as a scalable and extensible path toward stronger LLM agents.
\end{enumerate}

\section{Methodology}

\subsection{Vanilla Formulation}
Given a question $q$, an agent generates a multi-turn interaction trajectory denoted as
\begin{equation}
    \tau := (a_1, o_1, a_2, o_2, \ldots, a_T, o_T),
\end{equation}
where $a_i$ is the LLM output at step $i$ (including \textit{reasoning} and \textit{tool call}), and $o_i$ is the corresponding tool-call result.
The vanilla ReAct-style agent \citep{Yao2022ReAct} models the interaction as following,
\begin{equation}
   p^{\mathrm{ReAct}}_{\theta}(\tau \mid q) = \prod_{i\in[T]} \pi_{\theta}\big(a_i \mid q, (a_1,o_1,\ldots, a_{i-1}, o_{i-1})\big),
\end{equation}
which appends the entire interaction history to the context at each time of LLM generation. 
However, in long-horizon agentic tasks like deep research and agentic coding, $\tau$ can accumulate rapidly due to extensive interactions and become prohibitively long which exceeds the working context limit.
Also, when the context is expanding, the reasoning and instruction following capability of the model may drop, posing further challenges for the agent to complete the long-horizon task.

\subsection{Our Method: Context Folding}
To address the challenge, we introduce context folding, a mechanism that allows the agent to \textit{actively} manage its working context via \textit{branching and folding}. 
Specifically, we design two tools that the agent can call for context management.
Starting from a main thread to solve question $q$, it can:
\begin{enumerate}[label=(\roman*),leftmargin=20pt]
    \item \code{\textcolor{red}{branch}(description, prompt)}: \textit{branch from main thread to use a separate working context to complete a sub-task $q'$ for solving $q$}. 
    Here \code{description} is a brief summary of the sub-task, and \code{prompt} is a detailed instruction for this branch. 
    The tool returns a template message indicating that the branch was created. 
    \item \code{\textcolor{blue}{return}(message)}: \textit{fold the context generated in this branch and return to the main thread}. The \code{message} describes the outcome of this branch.
    Upon calling this tool, the agent context then switches back to the main thread, appended with the templated \code{message} from the branch.
\end{enumerate}
With these two tools, the agent can actively manage its context by (i) branching a separate working context to solve an independent sub-task, and (ii) folding the intermediate steps in the branch, and resuming back to the main thread by appending only the result of the branch.
To put it formal, the context-folding agent is modeled as following,
\begin{align}
    p^{\text{Context Fold}}_{\theta}(\tau \mid q):= \prod_{i\in[T]} \pi_{\theta}\big(a_i\mid q, \mathcal{F}(\tau_{<i})\big).\label{eq: context fold model}
\end{align}
Here $\tau_{<i} = (a_1,o_1,\ldots, a_{i-1}, o_{i-1})$ denotes the complete history of all the action-observation pairs before step $i$, 
$\mathcal{F}$ is the context manager that folds the interaction history between \code{\textcolor{red}{branch}} and \code{\textcolor{blue}{return}} tool calls.
We illustrate the process using the following example, where the context manager folds all the action-observation pairs in previous branches:
\begin{align}
\mathcal{F}\big(
a_1, o_1,
\textcolor{red}{a_2},
\underbrace{\textcolor{red}{o_2}, a_3, o_3, \textcolor{blue}{a_4}}_{\displaystyle{\small\text{branch 1}}},
\textcolor{blue}{o_4},
\textcolor{red}{a_5},&
\underbrace{\textcolor{red}{o_5}, a_6, o_6, a_7,o_7,\textcolor{blue}{a_8}}_{\displaystyle{\small\text{branch 2}}},
\textcolor{blue}{o_8}, a_9, o_9, \textcolor{red}{a_{10}}, o_{10}
\big) \\
& \to
\big(
a_1, o_1, \textcolor{red}{a_2}, \textcolor{blue}{o_4}, \textcolor{red}{a_5}, \textcolor{blue}{o_8}, a_9, o_9, \textcolor{red}{a_{10}}, o_{10}
\big),
\end{align}
so the segments between $\textcolor{red}{a_2}$ and $\textcolor{blue}{a_4}$ and between $\textcolor{red}{a_5}$ and $\textcolor{blue}{a_8}$ are folded.
 
\textbf{Inference efficiency.} 
During inference, the agent manages a context KV-cache: when \textcolor{blue}{\code{return}} action is called, it rolls back the KV-cache to the corresponding \textcolor{red}{\code{branch}} position, where the context prefix matches that before calling the \textcolor{red}{\code{branch}} action.
This makes our context folding approach efficient in terms of inference.

\textbf{Instantiation: plan-execution.}
To instantiate context folding for long-horizon tasks, we adopt a \textit{plan–execution} framework, where the agent alternates between two states:  
\textit{(i) Planning State}: The agent performs high-level reasoning in the main thread, decomposes the task, and decides when to initiate a branch for a sub-task. 
In this state, token-intensive tool use is discouraged to keep the main context focused on high-level strategies.
\textit{(ii) Execution State}: The agent operates within an active branch to complete its assigned sub-task. To maintain a clear structure and prevent nested complexity, creating new branches is disabled while in execution state.

\begin{figure}
    \centering
    \includegraphics[width=1.0\linewidth]{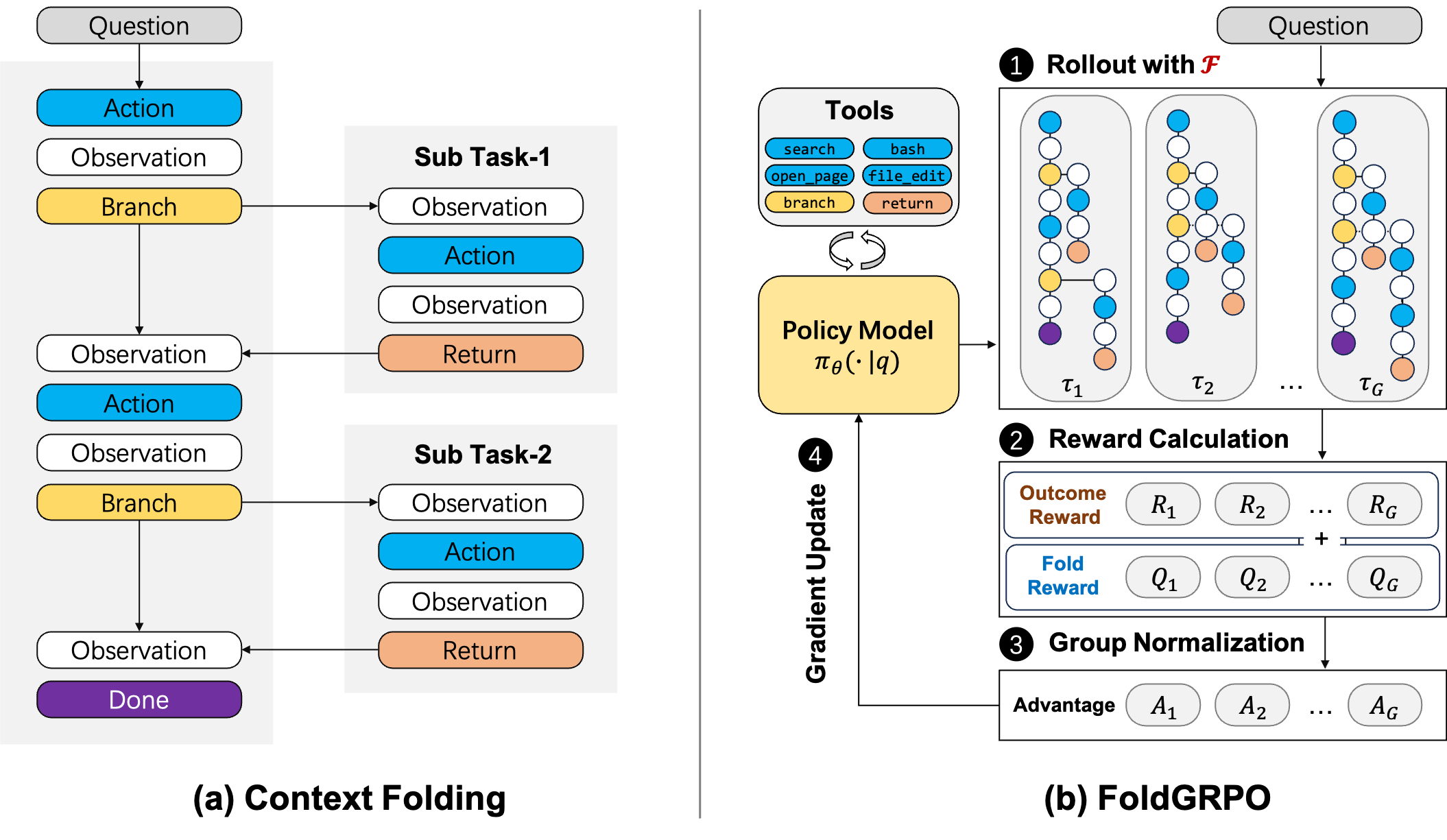}
    \vspace{-0.5cm}
    \caption{\textbf{(a) Context Folding:} a mechanism that enables the agent to actively manage its context through branching and return. \textbf{(b) FoldGRPO:} end-to-end optimization of context folding agent.}
    \label{fig:model-main}
\end{figure}

\subsection{FoldGRPO: End-to-End RL for Context-Folding Agent}
To optimize the context folding agent, in this section, we introduce an end-to-end RL training framework, namely, \underline{Fold}ed-context \underline{G}roup \underline{R}elative \underline{P}olicy \underline{O}ptimization (FoldGRPO). 
FoldGRPO jointly optimizes the entire interaction trajectory including the main thread and those sub-task branches, while it folds the rollout history according to the context folding modeling \eqref{eq: context fold model} to maintain \textit{a compact working context} during training.
Moreover, FoldGRPO features a novel \textit{process reward design} to efficiently guide the training of the branching behavior of the agent. 
We first introduce the overall algorithm design in Section~\ref{subsubsec: overall} and we present the process reward design in Section~\ref{subsubsec: prm}.

\subsubsection{Overall Algorithm Design}\label{subsubsec: overall}
In each training step of FoldGRPO, for task $q$ from training dataset $\mathcal{D}$, $G$ trajectories $(\tau_1, \tau_2, \cdots, \tau_G)$ are sampled from the old policy $\pi_{\text{old}}$ according to the context folding model \eqref{eq: context fold model}.
Each complete trajectory, e.g., $\tau_i = (a_{i,1}, o_{i,1}, \cdots,a_{i,T},o_{i,T})$, is a sequence of tokens defined as $\tau_i = [\tau_{i,1}, \cdots, \tau_{i,|\tau_i|}]$.
Each trajectory $\tau_i$ has a final reward $R_i \in \{0, 1\}$, following the recipe of RL from verifiable rewards (RLVR).

\textbf{Learning objective.}
The learning objective of FoldGRPO is defined as:
\begin{equation}\small
\mathcal{J}_{\text{FoldGRPO}} = \mathbb{E}_{\substack{q\sim \mathcal{D},\\ \{\tau_i\}_{i=1}^{G}\sim \pi_{\text{old}}(\cdot|q)} }
\left[
\frac{1}{\sum_{i=1}^{G} |{\tau}_i|}
\sum_{i=1}^{G} \sum_{t=1}^{|{\tau}_i|}
\min\Big\{
r_{i,t}(\theta) \widehat{A}_{i,t},
\ \mathrm{clip}\big(r_{i,t}(\theta), 1-\epsilon_{\text{low}}, 1+\epsilon_{\text{high}}\big)\widehat{A}_{i,t}
\Big\}
\right],
\end{equation}
where the importance sampling ratio and the group relative advantage estimator \citep{shao2024deepseekmath} are given by
\begin{equation}
r_{i,t}(\theta) =
\frac{\pi_{\theta}({\tau}_{i,t}\mid q, \mathcal{\textcolor{red}{F}}({\tau}_{i, <t}))}
     {\pi_{\theta_{\text{old}}}({\tau}_{i,t}\mid q, \mathcal{\textcolor{red}{F}}({\tau}_{i, <t}))}\cdot
     \mathbf{1}_{\tau_{i,t}}^{\mathrm{LLM}},
\quad
\widehat{A}_{i,t} =
\frac{\mathrm{clip}(R_i + \textcolor{red}{Q_{i, t}}, 0, 1) - \mathrm{mean}(\{R_i\}_{i=1}^{G})}
     {\mathrm{std}(\{R_i\}_{i=1}^{G})}.
\end{equation}
Here $\mathbf{1}_{\tau_{i,t}}^{\mathrm{LLM}}$ ensures that only those LLM generated tokens are optimized and the tokens from tool observations are masked.
In the following, we explain two key features of FoldGRPO highlighted in red.

\begin{enumerate}[label=(\roman*),leftmargin=20pt]
\item \textbf{Context folding.}
Unlike vanilla multi-turn LLM RL algorithms that append the entire interaction history to context when optimizing the policy, FoldGRPO
applies context manager $\mathcal{F}(\cdot)$ to the history ${\tau}_{i,<t}$ which folds the context for token ${\tau}_{i,t}$ based on the branch-return actions.

\item \textbf{Process reward signal.}
In the calculation of advantage $\widehat{A}_{i,t}$, a token-level process reward $Q_{i,t}$ is added to regularize the model’s branch-return behavior, which is detailed in the next section.
\end{enumerate}

\subsubsection{Process Reward Design}\label{subsubsec: prm}

In RLVR, the agent is typically optimized through a standard binary \textit{outcome reward} based on task success or failure.
However, we empirically observe that this sparse reward signal is insufficient for learning effective context folding. 
Specifically, two critical failure modes emerge:
(i) The agent fails to plan strategically, leaving token-intensive operations unfolded in the main context, which quickly exhausts the available token budget.
(ii) The agent struggles with proper branch management, often failing to return from a sub-branch after a sub-task is completed and instead continuing the subsequent work within that same branch.
To effectively optimize the folding agent, we introduce token-level process rewards separately to main-trajectory tokens and branch-trajectory tokens.

\noindent \textbf{Unfolded token penalty.} When total context length of the main thread exceeds 50\% of the working context limit, we apply $Q_{i,t}=-1$ to all the tokens in the main thread, except those tokens in the turns that create a branch. This penalizes the agent for performing token-heavy actions outside a branch in the main thread, and encourages the agent to perform those actions in separate branches.

\noindent \textbf{Out-scope penalty.} For each branch, we employ GPT-5-nano to judge — based on the branch prompt and the returned message — whether the agent has conducted actions outside the specified sub-tasks. If so, we apply $Q_{i,t}=-0.2$ to all the tokens in this branch to penalize such out of scope behavior.
This encourages the agent to only perform the exact sub-task given to the current branch.

\textbf{Failure penalty.} We apply $Q_{i,t}=-1$ to all the tokens in a failed tool call turn. 
In all other cases, $Q_{i,t}=0$.

\subsection{How does Context Folding Connect to Other Methods?}
\textbf{Relationship to multi-agent systems.}
Context folding can be interpreted as a specific formulation of a general multi-agent system, where the main agent delegates sub-tasks to sub-agents.
Compared to popular multi-agent systems~\citep{Hadfield2025}, our design differs in the following ways:
(i) Context folding does not adopt predefined sub-agents; instead, sub-agents are created by the main agent on the fly;
(ii) All the agents share the same context prefix, making it KV-cache friendly,
(iii) The main and the sub agents interleave rather than operating in parallel.

\textbf{Relationship to context-summarization-based method.}
Compared with heuristic summarization-based context management~\citep{Yu2025MemAgentRL,OpenAI2025a}, which discards details at arbitrary points, context folding can be viewed as a learnable summarization mechanism aligned with sub-task boundaries. 
This ensures that reasoning is preserved during execution and is only compacted once its utility is realized.

\section{Experiment}

\subsection{Datasets}
We conduct experiment on two representative long-horizon agent tasks: deep research, and agentic software engineering:

\noindent \textbf{Deep Research.}
We use BrowseComp-Plus (BC-Plus)~\citep{Chen2025BrowseCompPlusAM}, which supplements the original BrowseComp data with a verified corpus. 
We use Qwen3-Embed-8B as the retriever.
Since the quality of training data is crucial for the BrowseComp task but existing datasets are typically not open-sourced~\citep{Qiao2025WebResearcherUU,Li2025WebSailorV2BT}, we split BrowseComp-Plus into training and evaluation sets to decouple the effect of data distribution. 
Our split includes 680 instances for training and 150 for evaluation. 
For deep research, the tools are \code{search(query, topk)} and \code{open\_page(url)}, and the reward is based on official LLM-based judger~\citep{Chen2025BrowseCompPlusAM}.

\noindent \textbf{Agentic SWE.} 
We use SWE-Bench Verified (SWEB-V)~\citep{Jimenez2023SWEbenchCL} as the evaluation set. 
To collect training data, we roll out the baseline agent\footnote{Seed-OSS-36B-Instruct with OpenHands and a response length of 65,536.} eight times on a subset of the open-source datasets SWE-Gym~\citep{Pan2024TrainingSE} and SWE-Rebench~\citep{Badertdinov2025SWErebenchAA}, and retain the instances where the success rate is between 0 and $87.5\%$, resulting in 740 instances. 
In SWE, the tools are \code{execute\_bash}, \code{str\_replace\_editor}, and \code{think}~\citep{Wang2024OpenHandsAO}, and the reward is based on running unit test in instance-specific sandbox environment.

We classify test instances for both tasks into three difficulty levels: \textit{easy}, \textit{medium}, and \textit{hard}. For BrowseComp-Plus, classification is determined by running a ReAct agent 8 times on each instance. An instance is labeled \textit{easy} if its acc@8 score is $\geq87.5\%$, \textit{hard} if its score is 0\%, and \textit{medium} otherwise, resulting in 50 instances for each level. For SWE-Bench Verified, classification is based on the original dataset's time-to-resolve metric: \textit{easy} ($\le$15 min, 194 instances), \textit{medium} (15 min -- 1 hour, 261 instances), and \textit{hard} ($\ge$1 hour, 45 instances).

See Appendix \ref{sec:pe} for the details of system prompt of each datasets.

\subsection{Implementation}
We use Seed-OSS-36B-Instruct\footnote{\url{https://huggingface.co/ByteDance-Seed/Seed-OSS-36B-Instruct}} as the base LLM and conduct RL training on it.  
For RL training, we build on VeRL and set the rollout batch size to 32, group size to 8, ppo batch size of 128, learning rate to $1 \times 10^{-6}$, no KL term, clip high to 0.28, and clip low to 0.2.  
We employ asynchronous rollout with a maximum off-policy step of 5.  
During training, we implement the context folding operation $\mathcal{F}$ by constructing separate causally conditioned contexts for each branch to improve training efficiency (See Appendix~\ref{sec:algo-impl} for more details.). 
We train model for 50 steps (about 2 epochs).
For the fold agent, we set the LLM maximum context length to 32,768.
We allow up to 10 branches, resulting in a theoretical maximum of 327,680 tokens.
During inference we employ greedy decoding (i.e, temperature = 0).

\input{table/main}

\subsection{Baselines}
We compare against the following baselines:

\noindent \textbf{ReAct Agent}~\citep{Yao2022ReActSR}, which keeps all context. We consider different context lengths for comparison: (a) \textit{short-context}, which has 32,768 tokens, equivalent to our context length; (b) \textit{medium-context}, which has intermediate lengths, e.g., 65,536 and 131,072; (c) \textit{long-context}, which has 327,680 tokens, equivalent to our maximum total token cost.

\noindent  \textbf{Summary Agent}~\citep{Yu2025MemAgentRL,Wu2025ReSumUL}, which invokes a summary when the context is full. We set the maximum context length to 32,768 and allow for 10 summary session for a fair comparison.

For both two baselines, we employ the same base model (i.e., Seed-OSS-36B-Instruct), data, infrastructure, and hyperparameters for RL training.
In addition to these directly comparable baselines, we compare our method against previous closed-source and open-source systems on both tasks, including GPT-5, GPT-4.1, DeepSeek-V3.1 (2509), GLM-4.5-Air, and Qwen3-235B-A22B-Instruct-2507.

\begin{figure}[t!]
    \centering
    \includegraphics[width=1.0\linewidth]{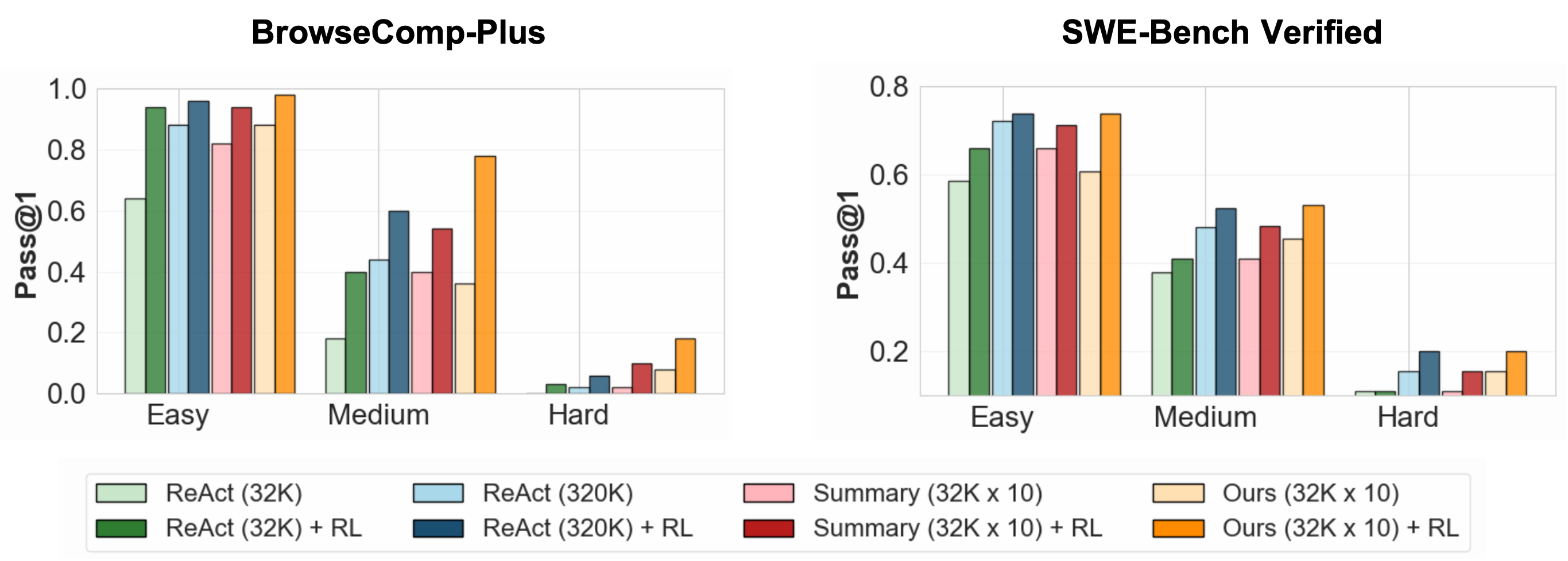}
    \vspace{-0.5cm}
    \caption{Agent performance on different data difficulty group. RL training yields consistent performance gains across easy, medium, and hard instances.}
    \label{fig:diff}
\end{figure}
\begin{figure}[t!]
    \centering
    \includegraphics[width=0.98\linewidth]{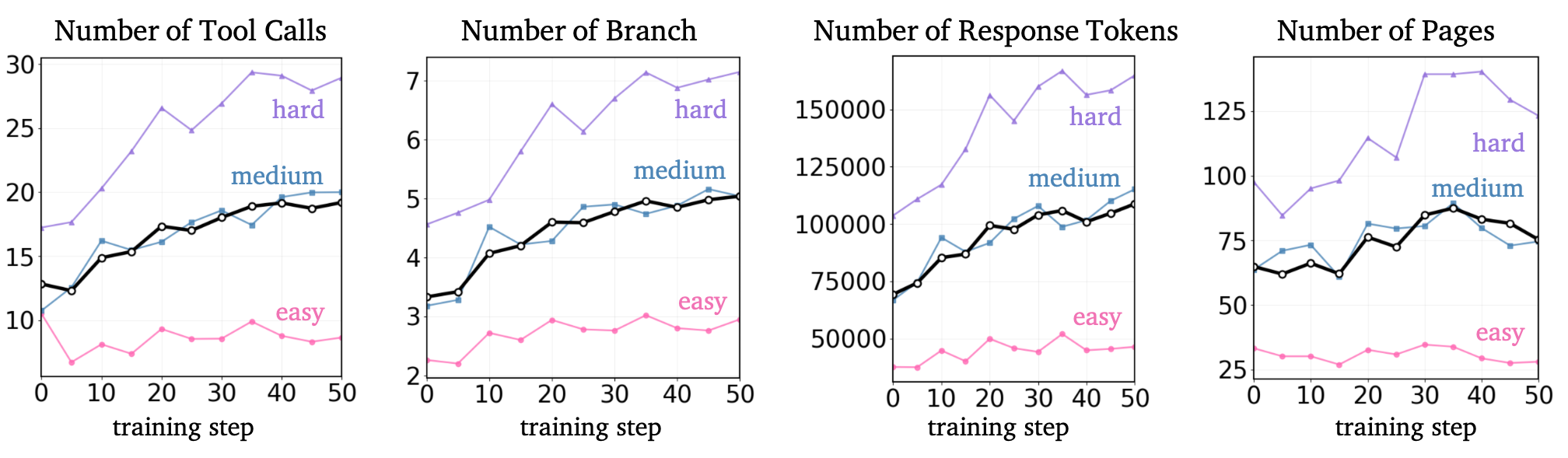}
    \vspace{-0.2cm}
    \caption{With RL training, we observe an increase in the number of tool calls, branching behavior, total number of tokens, and the number of searched pages.}
    \label{fig:tool_call}
\end{figure}

\section{Experimental Results}

\subsection{Main Results}
Table~\ref{tab:main} summarizes our main results on the BrowseComp-Plus and SWE-Bench Verified datasets. Our findings highlight the critical role of reinforcement learning in unlocking the capabilities of context folding.

Initially, without performing RL, the context folding agent already surpasses the 32K-context ReAct and context summarization baselines, though it does not yet match the performance of the long-context ReAct agent. 
After RL training, our agent's performance improves significantly, with a $pass@1$ of \textbf{0.620 on BrowseComp-Plus (+20\%)} and \textbf{0.580 on SWE-Bench Verified (+8.8\%)}. Our agent not only outperforms all baselines, including the long-context ReAct agent with same 327K max length, but also achieves performance comparable to agents built on much larger 100B+ parameter models.

Further analysis reveals two key insights. 
First, an ablation study confirms that our proposed \textbf{FoldGRPO is crucial}, yielding significantly better performance than the baseline GRPO algorithm (eg, +7.7\% on BrowseComp and +1.6\% on SWE-Bench). Second, the performance gains correlate with an increased frequency of tool calls, which RL training further encourages. This suggests our framework enables the agent to conduct a more thorough exploration of the environment to discover more robust solutions.

\subsection{Performance by Task Difficulty}
Figure~\ref{fig:diff} breaks down agent performance by task difficulty, comparing scores before and after reinforcement learning. The results clearly show that RL training yields consistent performance gains across easy, medium, and hard instances. Most notably, the improvements are significantly larger for the medium and hard subsets. This finding underscores our agent's enhanced capability to handle complex problems that require more sophisticated long-context management.

Figure~\ref{fig:tool_call} shows the agent’s learning dynamics during RL training on BrowseComp-Plus. As training progresses, the agent steadily increases its tool calls, branch creation, response tokens, and number of pages searched. This growth is most pronounced on harder instances. For example, on the hard subset, response length rises from about 100K to over 160K tokens. These results show that the agent learns to allocate more interaction and computation to complex problems, adopting a more adaptive and effective problem-solving strategy.

\subsection{Ablation of RL Algorithm}
To understand how our proposed FoldGRPO shapes agent behavior, we analyze the key statistics in Table~\ref{tab:action}. These metrics include the task completion rate within the context limit (Finish), main trajectory length (Main Len), the accuracy of sub-trajectories staying on-topic (Scope), and the number of branches created (\# Branch).
We can see that, training with a standard GRPO baseline produces poor behaviors: agents show a lower Finish rate, generate overly long trajectories, and lose focus in sub-tasks, reflected in reduced Scope accuracy. This indicates a failure to manage context effectively.

By contrast, our FoldGRPO corrects these issues. It encourages focused branching, sharply boosting both Scope accuracy and Finish rate. Most notably, it cuts the main trajectory to about 8K tokens while processing over 100K in total—achieving over \textbf{90\% context compression} and demonstrating the agent’s ability to condense long interactions into a compact, useful history.

\input{table/ablation}

\begin{figure}[t!]
    \centering
    \includegraphics[width=0.75\linewidth]{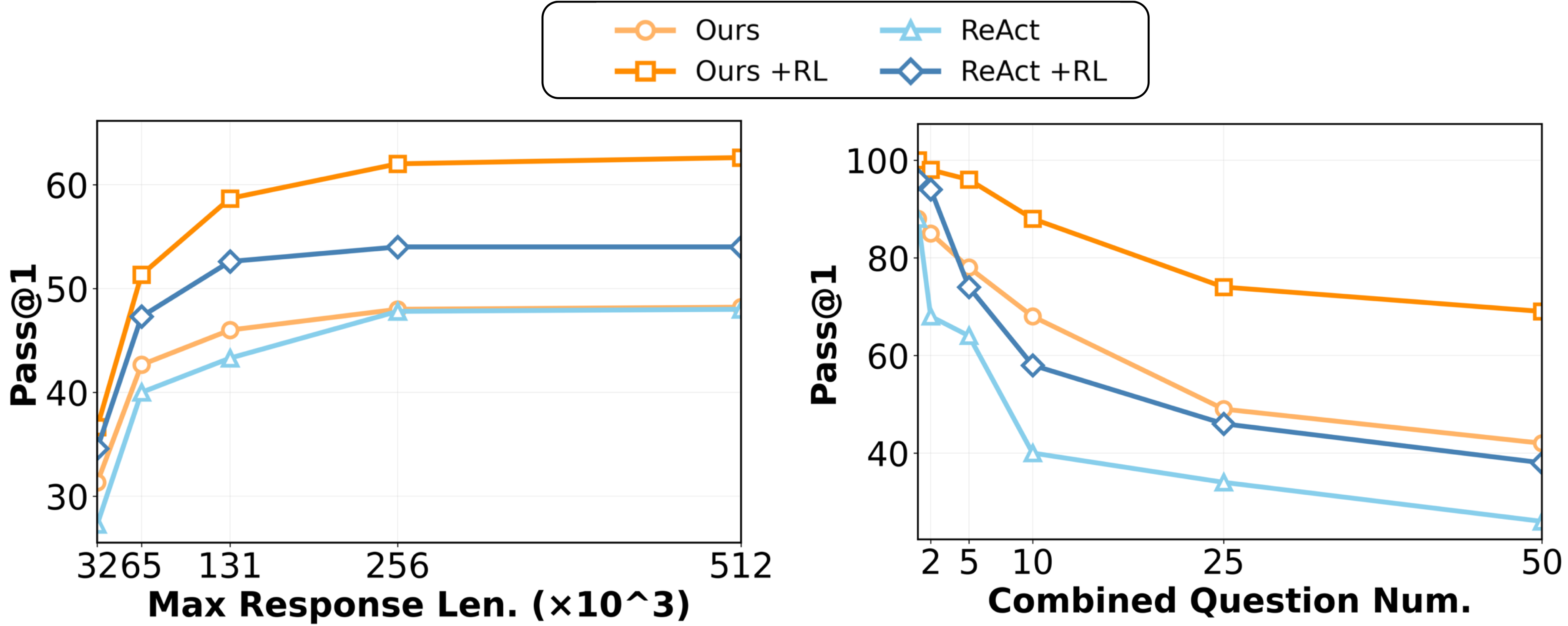}
    \vspace{-0.2cm}
    \caption{\textbf{Left:} Pass@1 vs. agent max context length. \textbf{Right:} Pass@1 vs. number of combined questions. Multiple easy questions are combined into a single harder question to increase problem complexity; a higher number of combined questions indicates more required actions and a longer context to answer them correctly. See Section~\ref{sec:multiq} for details.}
    \label{fig:multiq}
\end{figure}

\subsection{Performance by Context Length}

\subsubsection{Effect of Context Length}
To examine how performance scales with context length, we evaluated our method on BrowseComp while varying the number of branches from 0 to 16. As shown in Figure~\ref{fig:multiq} (left), our method consistently surpasses ReAct, and reinforcement learning provides further gains. However, performance plateaus beyond 320K tokens because most task instances are already completed, and additional context provides limited benefit.

\subsubsection{Effect of Task Complexity}\label{sec:multiq}
\begin{figure}[h]
\centering
\includegraphics[width=1\columnwidth]{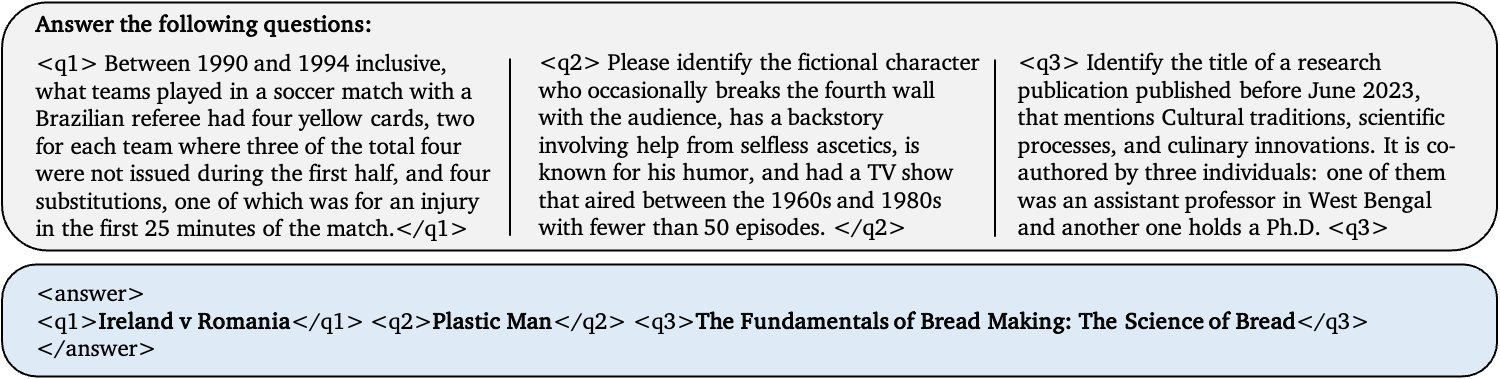}
\caption{An example of the model's input and output for the combined-questions experiment described in Section~\ref{sec:multiq}. In this example, 3 easy questions are combined to form a harder question.}
\label{fig:multiq-exp}
\end{figure}

Following \citet{Zhou2025MEM1LT}, we increase task complexity by combining multiple questions into a single compound query that the agent must answer \textit{in one session} (see Figure~\ref{fig:multiq-exp} for an example). Figure~\ref{fig:multiq} (right) shows the results for tasks with 1 to 50 combined questions. For this setting, we allow unlimited branching and set the context limit for ReAct to 1M tokens. As task complexity increases, the benefit of context folding becomes more apparent, demonstrating strong length generalization. Notably, although our agent was trained on tasks requiring at most 10 branches, it adaptively uses an average of 32.6 branches to solve tasks with 50 questions.

\subsection{Further Analysis}
\subsubsection{Case Study}
\begin{figure}[t!]
\centering\includegraphics[width=1.0\linewidth]{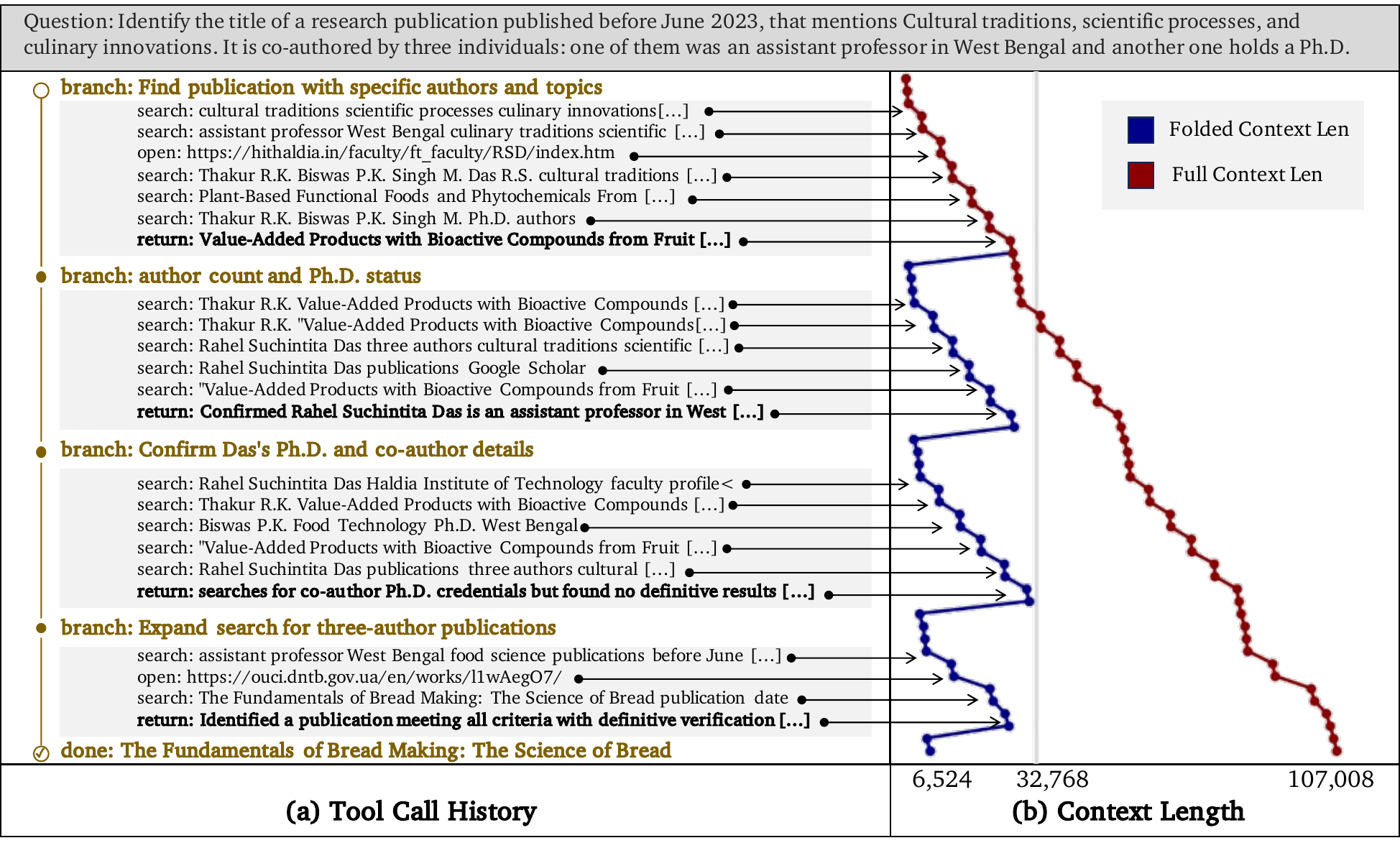}
\vspace{-0.6cm}
\caption{Example of an agent's tool call history and context length. on BrowseComp-Plus.}\label{fig:case}
\end{figure}

Figure~\ref{fig:case} shows qualitative examples of our context folding agent on BrowseComp-Plus. Given a query about finding a research publication with specific conditions, the agent first explores the high-level topic and identifies a candidate. It then searches to verify conditions, gaining key insights but failing to confirm all requirements. Next, it expands the search scope and finds the correct answer. In this process, 4 branches compress the full 107K-token context to just 6K.

\subsubsection{Training Speed}
\begin{figure}[t!]
    \centering
    \includegraphics[width=0.4\columnwidth]{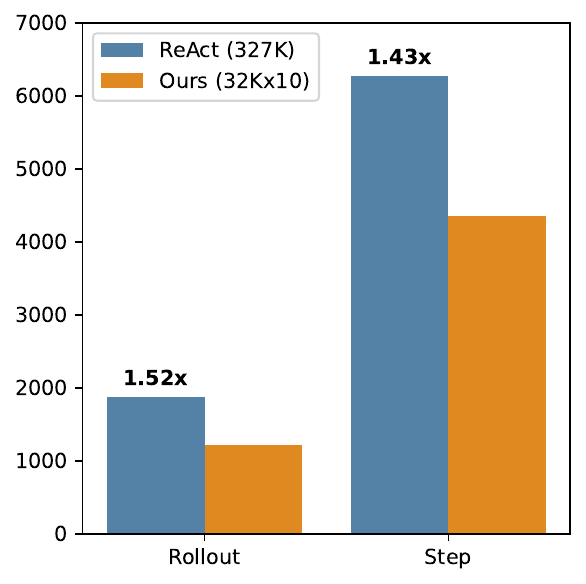}
    \vspace{-0.2cm}
    \caption{Training time cost. The figure shows the stepwise average time for rollout and for each training step.}
    \label{fig:speed}
\end{figure}

Figure~\ref{fig:speed} shows the stepwise average time for rollout and for each training step. We observe that the 327K ReAct model requires a longer training time per step.
Note that we employ async rollout (Appendix~\ref{sec:async-rollout}), and the rollout time shown here measures only the main thread’s time cost on rollout.

\subsubsection{Parallel Branching}
Whether the folding agent can benefit from parallel branching — i.e., creating multiple sub-branches that run simultaneously — remains an open question.
We experimented on BrowseComp-Plus by training an agent that utilizes parallel branching under the same setup as the single-branch agent.
The parallel-branch version achieved a 0.6133 Pass@1 on BrowseComp-Plus, outperforming the baseline but performing similarly to the single-branch version.
Moreover, after training, the parallel-branch agent created about 2.3 parallel branches on average and read more web pages (110 vs. ~80 for the single-branch version).
However, it did not achieve a higher score, possibly because the task characteristics are more depth-first in nature. Other tasks with a breadth-first structure (eg WideSearch~\citep{Wong2025WideSearchBA}) may be more promising for studying parallelism in LLM agents.

\section{Related Work}
The rapid evolution of LLM agents is driven by a push toward greater autonomy in complex, long-horizon tasks~\citep{Jimenez2023SWEbenchCL,openai2025chatgptAgent,Zhou2023WebArenaAR,METR2025a,Li2025TowardsCA}. Built on agentic architectures that integrate planning, memory, and tool use~\citep{Wang2023ASO}, research has advanced from simple sequential reasoning to dynamic, multi-path strategies for exploration and problem-solving~\citep{Yao2023TreeOT,Besta2023GraphOT,Huang2022LanguageMA,Sun2025COBenchBL}. Yet this progress has revealed a key bottleneck: the finite and costly nature of an agent’s working context~\citep{Yao2022ReAct,OpenHands2025}.

Context management strategies fall into two main paradigms: context summarization, where agents offload and retrieve information from external memory stores~\citep{Wang2023VoyagerAO,Tang2025AgentKL,Yu2025MemAgentRL,Wu2025ReSumUL,Zhou2025MEM1LT}, and multi-agent collaboration, where tasks are divided among specialized agents with focused contexts~\citep{Zhao2024LongAgentSL,Zhang2024ChainOA,Anthropic2025c,Wong2025WideSearchBA}.
Both paradigms frame context management as an architectural or retrieval problem, leaving a gap for an integrated approach where it becomes a learned cognitive skill rather than an external feature.

Reinforcement learning (RL) effectively grounds agents through environmental or human feedback~\citep{Zhang2025TheLO,Qiao2025WebResearcherUU}, but has focused mainly on extrinsic task success~\citep{DeepSeekAI2025DeepSeekR1IR}. 
The training of intrinsic skills—such as how an agent manages its own working memory—remains a underexplored research area. Our work contributes to this emerging frontier by framing context management as a learnable skill and using process-level rewards to teach it directly.

\section{Conclusions and Future Work}
In this paper, we introduced \textbf{context folding}, an agentic mechanism for managing long-horizon trajectories by selectively folding ephemeral sub-trajectories while preserving only essential decision-relevant information. Coupled with our reinforcement learning framework, context folding enables efficient credit assignment across tree-structured trajectories and achieves significant improvements in long-horizon coding and deep-research tasks. Empirical results on two long-context tasks demonstrate that folding allows agents to match or exceed the performance of baselines with larger context windows, while improving efficiency and stability relative to summary-based condensation. 
Several promising future directions include multi-layer context folding, which develops hierarchical folding strategies where folds themselves can be further folded for deeper compression.

\section*{Acknowledgments}
The authors would thank Weihua Du, Guanghao Ye, Joey Hong, Bowen Xiao, Ting-Han Fan, Lingfeng Shen for valuable discussions and feedback during the preparation of this work.

%% file: table/main.tex
\begin{table}[t!]
\centering\small
\setlength{\tabcolsep}{7pt}
\begin{tabular}{lll|cc|cc}
\toprule
 &  &  & \multicolumn{2}{c|}{\textbf{BrowseComp-Plus}} & \multicolumn{2}{c}{\textbf{SWE-Bench Verified}} \\
\textbf{Model} & \textbf{Peak Length} & \textbf{Max \#Token} & \textbf{Pass@1} & \textbf{Tool Calls} & \textbf{Pass@1} & \textbf{Tool Calls} \\
\midrule
\rowcolor{gray!10}\multicolumn{7}{c}{\textbf{ReAct Agent with 100B+ LLM}} \\
GPT-5   & 327K & 327K & 0.793 & 14.2 & 0.718   & 42.6 \\
GPT-4.1  & 327K & 327K & 0.640 & 5.6 & 0.486   & 28.7 \\
DeepSeek-V3.1  & 327K & 327K & 0.613 & 10.6 & 0.610   & 53.2 \\
GLM-4.5-Air  & 327K & 327K   & 0.566    & 11.1 & 0.576   & 51.2 \\
Qwen3-235B-A22B  & 327K & 327K & 0.560 & 12.8 & 0.344   & 32.1 \\

\midrule
\rowcolor{gray!10}\multicolumn{7}{c}{\textbf{ReAct Agent}} \\
Seed-OSS-36B   & 32K & 32K  & 0.286 \sred{(-19.2)} & 3.8 & 0.436 \sred{(-11.6)} & 25.8\\
\quad + RL (GRPO)              & 32K & 32K  & 0.446 \sred{(-3.2)\phantom{0}} & 5.5 & 0.480 \sred{(-7.2)\phantom{0}}   & 27.8 \\
Seed-OSS-36B$^\psi$     & 327K & 327K & 0.478 {\tiny (+0.0)} & 10.8 & 0.552 {\tiny (+0.0)\phantom{0}} & 49.5 \\
\quad + RL (GRPO)            & 327K & 327K & 0.540 \sgreen{(+6.2)} & 10.2 & 0.574 \sgreen{(+2.2)\phantom{0}} & 55.4 \\

\midrule
\rowcolor{gray!10}\multicolumn{7}{c}{\textbf{Summary Agent}} \\
Seed-OSS-36B     & 32K & 32K $\times$ 10 & 0.386 \sred{(-9.2)\phantom{0}} & 17.4 & 0.488 \sred{(-6.4)\phantom{0}} & 77.0 \\
\quad + RL (GRPO)  & 32K & 32K $\times$ 10 & 0.527 \sgreen{(+4.9)} & 18.0 & 0.550 \sred{(-0.2)\phantom{0}} & 74.9 \\

\midrule
\rowcolor{gray!10}\multicolumn{7}{c}{\textbf{Folding Agent (Ours)}} \\
Seed-OSS-36B     & 32K & 32K $\times$ 10 & 0.420 \sred{(-5.8)\phantom{0}} & 12.9 & 0.492 \sred{(-6.0)\phantom{0}} & 72.8 \\
\quad + RL (GRPO)             & 32K & 32K $\times$ 10 & 0.567 \sgreen{(+8.9)} & 16.0 & 0.564 \sgreen{(+1.2)} & 79.5 \\
\quad \textbf{+ RL (FoldGRPO)}        & 32K & 32K $\times$ 10  & \textbf{0.620} \sgreen{(+14.2)} & 19.2 & \textbf{0.580} \sgreen{(+2.8)} & 96.5 \\
\bottomrule
\end{tabular}
\caption{\textbf{Performance on BrowseComp-Plus (N=150) and SWE-Bench Verified (N=500).} Boldface indicates the best-performing 36B models. Numbers in parentheses indicate improvement or reduction compared to 327K ReAct agent Seed-OSS-36B baseline$^\psi$. }
\label{tab:main}
\end{table}

%% file: table/ablation.tex
\begin{table}[t!]
\centering\small
\setlength{\tabcolsep}{3pt}
\begin{tabular}{l|cccc|cccc}
\toprule
& \multicolumn{4}{c|}{\textbf{BrowseComp-Plus}} & \multicolumn{4}{c}{\textbf{SWE-Bench Verified}} \\
 & Finish & Main Len & Scope & \# Branch & Finish & Main Len & Scope & \# Branch \\
 
\midrule
Folding Agent (Seed-OSS-36B) & 0.806 & 12,195 & 0.774 & 3.51 & 0.781 & 47,475 & 0.473  & 3.05\\
\quad + RL (GRPO) & 0.738 & 22,285 & 0.762 & 3.88 & 0.612 & 48,908 & 0.419 & 3.80\\
\quad + RL (FoldGRPO) & 0.935 & 7,752 & 0.895 & 4.98 & 0.962 & 8,885 & 0.754 & 5.90\\
\bottomrule

\end{tabular}
\vspace{-0.5em}
\caption{\textbf{Model behavior statistics of different optimization methods.} FoldGRPO encourages focused branching and condensed main context, boosting both scope accuracy and finish rate.}\label{tab:action}
\end{table}

%% file: appendix.tex


\input{prompt}

%% file: prompt.tex
\section{Algorithm Implementation}\label{sec:algo-impl}

\subsection{Multi-Trajectories Collection}

\begin{figure}[h]
\centering
\includegraphics[width=1\columnwidth]{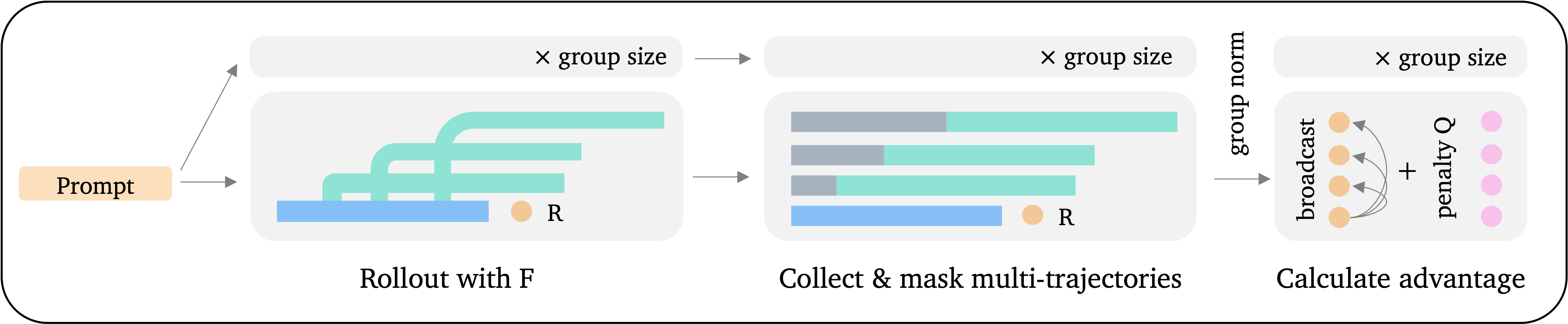}
\end{figure}

For practical implementation of model training, instead of concatenating all sub-trajectories into one sequence, we keep them as separate causally conditioned sequences, as shown above. Therefore, training with context folding is not directly compatible with existing training infrastructures (e.g., in Verl).

\subsection{Asynchronous Long-Horizon Agent Rollout}\label{sec:async-rollout}

\begin{figure}[h]
\centering
\includegraphics[width=1\columnwidth]{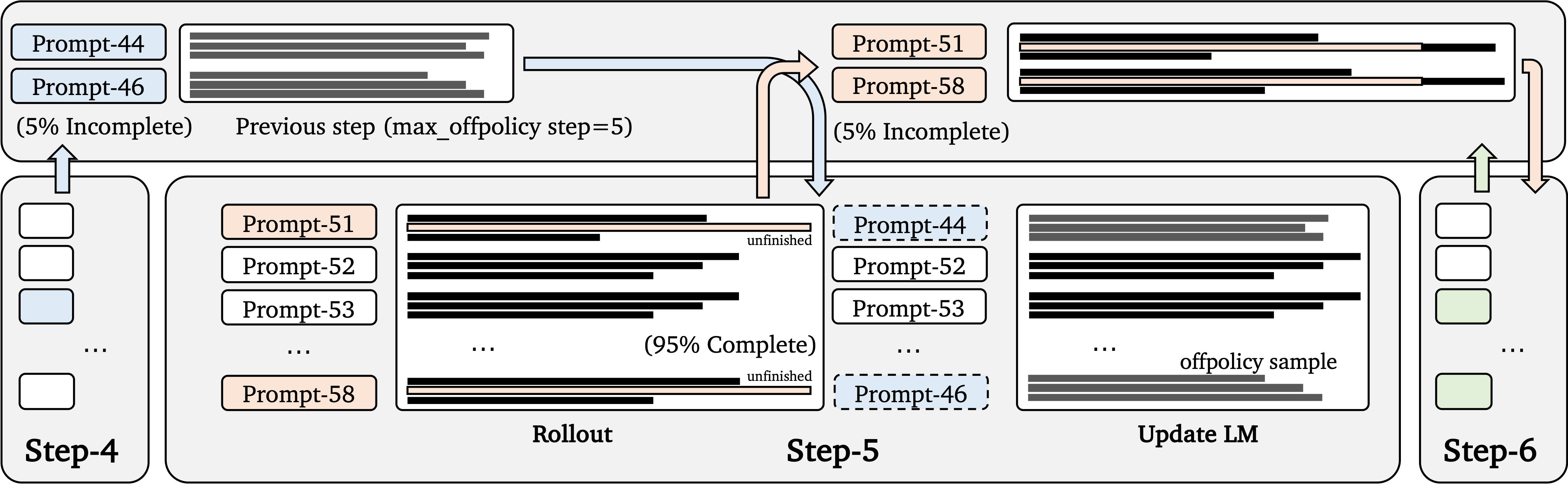}
\end{figure}

The rollout time of long-horizon agents is imbalanced, which causes a “bubble” in computation, where faster jobs wait for the longest one to finish. In our training setup, we mitigate this by adding an additional standalone rollout process: the main rollout process stops once it completes 95\% of the prompts (this hyperparameter is adjusted based on the GPU configuration), and the remaining jobs are handled by the standalone process. The data used for updating the LM include both (i) the 95\% of the current batch and (ii) the prompts from the previous step that were completed by the standalone rollout. Note that this part is off-policy; we set the maximum number of off-policy steps to 5 and observe no performance degradation compared to training on fully on-policy data.

\section{Prompt Engineering}
\label{sec:pe}

\subsection{BrowseComp-Plus Workflow}
Our prompt for BrowseComp-Plus is inspired by and modified from Claude Deep-Research. Using Seed-OSS-36B, we found that our system prompt achieves 0.478 accuracy, while the default system prompt in BrowseComp-Plus achieves only around 0.08.

{\footnotesize
\begin{Verbatim}[breaklines, breakanywhere]

Phase 1: Deconstruction & Strategy

1.  Deconstruct the Query:
    * Analyze the user's prompt to identify the core question(s).
    * Isolate key entities, concepts, and the relationships between them.
    * Explicitly list all constraints, conditions, and required data points (e.g., dates, quantities, specific names).
2.  Hypothesize & Brainstorm:
    * Based on your knowledge, brainstorm potential search vectors, keywords, synonyms, and related topics that could yield relevant information.
    * Consider multiple angles of inquiry to approach the problem.
3.  Verification Checklist:
    * Create a Verification Checklist based on the query's constraints and required data points. This checklist will be your guide throughout the process and used for final verification.

Phase 2: Iterative Research & Discovery

Tool Usage:
* Tools:
    * `search`: Use for broad discovery of sources and to get initial snippets.
    * `open_page`: Mandatory follow-up for any promising `search` result. Snippets are insufficient; you must analyze the full context of the source document.
* Query Strategy:
    * Start with moderately broad queries to map the information landscape. Narrow your focus as you learn more.
    * Do not repeat the exact same query. If a query fails, rephrase it or change your angle of attack.
    * Execute a minimum of 5 tool calls for simple queries and up to 50 tool calls for complex ones. Do not terminate prematurely.
* Post-Action Analysis: After every tool call, briefly summarize the key findings from the result, extract relevant facts, and explicitly state how this new information affects your next step in the OODA loop.
* <IMPORTANT>Never simulate tool call output<IMPORTANT>

You will execute your research plan using an iterative OODA loop (Observe, Orient, Decide, Act).

1.  Observe: Review all gathered information. Identify what is known and, more importantly, what knowledge gaps remain according to your research plan.
2.  Orient: Analyze the situation. Is the current line of inquiry effective? Are there new, more promising avenues? Refine your understanding of the topic based on the search results so far.
3.  Decide: Choose the single most effective next action. This could be a broader query to establish context, a highly specific query to find a key data point, or opening a promising URL.
4.  Act: Execute the chosen action using the available tools. After the action, return to Observe.

Phase 3: Synthesis & Analysis

* Continuous Synthesis: Throughout the research process, continuously integrate new information with existing knowledge. Build a coherent narrative and understanding of the topic.
* Triangulate Critical Data: For any crucial fact, number, date, or claim, you must seek to verify it across at least two independent, reliable sources. Note any discrepancies.
* Handle Dead Ends: If you are blocked, do not give up. Broaden your search scope, try alternative keywords, or research related contextual information to uncover new leads. Assume a discoverable answer exists and exhaust all reasonable avenues.
* Maintain a "Fact Sheet": Internally, keep a running list of key facts, figures, dates, and their supporting sources. This will be crucial for the final report.

Phase 4: Verification & Final Report Formulation

1.  Systematic Verification: Before writing the final answer, halt your research and review your Verification Checklist created in Phase 1. For each item on the checklist, confirm you have sufficient, well-supported evidence from the documents you have opened.
2.  Mandatory Re-research: If any checklist item is unconfirmed or the evidence is weak, it is mandatory to return to Phase 2 to conduct further targeted research. Do not formulate an answer based on incomplete information.
3.  Never give up, no matter how complex the query, you will not give up until you find the corresponding information.
4.  Construct the Final Report:
    * Once all checklist items are confidently verified, synthesize all gathered facts into a comprehensive and well-structured answer.
    * Directly answer the user's original query.
    * Ensure all claims, numbers, and key pieces of information in your report are clearly supported by the research you conducted.
\end{Verbatim}
}

\subsection{SWE-Bench Workflow}
Our prompt for SWE-Bench follows OpenHands.
{\footnotesize
\begin{Verbatim}[breaklines, breakanywhere]
Phase 1. READING: read the problem and reword it in clearer terms
   1.1 If there are code or config snippets. Express in words any best practices or conventions in them.
   1.2 Hightlight message errors, method names, variables, file names, stack traces, and technical details.
   1.3 Explain the problem in clear terms.
   1.4 Enumerate the steps to reproduce the problem.
   1.5 Hightlight any best practices to take into account when testing and fixing the issue

Phase 2. RUNNING: install and run the tests on the repository
   2.1 Follow the readme
   2.2 Install the environment and anything needed
   2.2 Iterate and figure out how to run the tests

Phase 3. EXPLORATION: find the files that are related to the problem and possible solutions
   3.1 Use `grep` to search for relevant methods, classes, keywords and error messages.
   3.2 Identify all files related to the problem statement.
   3.3 Propose the methods and files to fix the issue and explain why.
   3.4 From the possible file locations, select the most likely location to fix the issue.

Phase 4. TEST CREATION: before implementing any fix, create a script to reproduce and verify the issue.
   4.1 Look at existing test files in the repository to understand the test format/structure.
   4.2 Create a minimal reproduction script that reproduces the located issue.
   4.3 Run the reproduction script to confirm you are reproducing the issue.
   4.4 Adjust the reproduction script as necessary.

Phase 5. FIX ANALYSIS: state clearly the problem and how to fix it
   5.1 State clearly what the problem is.
   5.2 State clearly where the problem is located.
   5.3 State clearly how the test reproduces the issue.
   5.4 State clearly the best practices to take into account in the fix.
   5.5 State clearly how to fix the problem.

Phase 6. FIX IMPLEMENTATION: Edit the source code to implement your chosen solution.
   6.1 Make minimal, focused changes to fix the issue.

Phase 7. VERIFICATION: Test your implementation thoroughly.
   7.1 Run your reproduction script to verify the fix works.
   7.2 Add edge cases to your test script to ensure comprehensive coverage.
   7.3 Run existing tests related to the modified code to ensure you haven't broken anything.

8. FINAL REVIEW: Carefully re-read the problem description and compare your changes with the base commit {{ instance.base_commit }}.
   8.1 Ensure you've fully addressed all requirements.
   8.2 Run any tests in the repository related to:
     8.2.1 The issue you are fixing
     8.2.2 The files you modified
     8.2.3 The functions you changed
   8.3 If any tests fail, revise your implementation until all tests pass
\end{Verbatim}
}

\section{Agent Scaffold}

\subsection{BrowseComp-Plus}

Following \citep{Chen2025BrowseCompPlusAM}, in BrowseComp-Plus the agent can use the following tools:
{\footnotesize
\begin{Verbatim}[breaklines, breakanywhere]
search = {
    'type': 'function',
    'function': {
        "name": "search",
        "description": "Performs a web search: supply a string 'query' and optional 'topk'. The tool retrieves the top 'topk' results (default 10) for the query, returning their docid, url, and document content (may be truncated based on token limits).",
        "parameters": {
            "type": "object",
            "properties": {
                "query": {
                    "type": "string",
                    "description": "The query string for the search."
                },
                "topk": {
                    "type": "integer",
                    "description": "Return the top k pages.",
                }
            },
            "required": [
                "query"
            ]
        }
    }
}
open_page = {
    'type': 'function',
    'function': {
        'name': 'open_page',
        'description': (
            "Open a page by docid or URL and return the complete content. "
            "Provide either 'docid' or 'url'; if both are provided, prefer 'docid'. "
            "The docid or URL must come from prior search tool results."
        ),
        'parameters': {
            'type': 'object',
            'properties': {
                'docid': {
                    'type': 'string',
                    'description': 'Document ID from search results to resolve and fetch.',
                },
                'url': {
                    'type': 'string',
                    'description': 'Absolute URL from search results to fetch.',
                },
            },
            'required': [],
        },
    },
}
finish = {
    'type': 'function',
    'function': {
        'name': 'finish',
        'description': """Return the final result when you have a definitive answer or cannot progress further. Provide a concise answer plus a brief, evidence-grounded explanation.""",
        'parameters': {
            'type': 'object',
            'properties': {
                'answer': {
                    'type': 'string',
                    'description': 'A succinct, final answer.',
                },
                'explanation': {
                    'type': 'string',
                    'description': 'A brief explanation for your final answer. For this section only, cite evidence documents inline by placing their docids in square brackets at the end of sentences (e.g., [20]). Do not include citations anywhere else.',
                },
                'confidence': {
                    'type': 'string',
                    'description': 'Confidence: your confidence score between 0% and 100% for your answer',
                },
            },
            'required': ['answer', 'explanation'],
        },
    },
}
\end{Verbatim}
}

Following \citet{Chen2025BrowseCompPlusAM}, the \code{search} tool retrieves the \code{topk} (default as 10) documents using Qwen3-Embed-8B from the BrowseComp-Plus corpus and displays the first 512 tokens. 
The \code{open\_page} tool fetches the full document, which is truncated to the first 4096 tokens. 
When the agent calls \code{finish}, the \code{answer} field is used for correctness evaluation.

The system prompt is as shown in \ref{sec:pe} and the user prompt is question and tool-use description.

\subsection{SWE-Bench}
In SWE-Bench, we follow OpenHands~\citep{OpenHands2025}, the agent can use the following tools:
{\footnotesize
\begin{Verbatim}[breaklines, breakanywhere]
execute_bash = {
    'type': 'function',
    'function': {
        'name': 'execute_bash',
        'description': """Execute a bash command in the terminal.
* Long running commands: For commands that may run indefinitely, it should be run in the background and the output should be redirected to a file, e.g. command = `python3 app.py > server.log 2>&1 &`.
* One command at a time: You can only execute one bash command at a time. If you need to run multiple commands sequentially, you can use `&&` or `;` to chain them together.
""",
        'parameters': {
            'type': 'object',
            'properties': {
                'command': {
                    'type': 'string',
                    'description': 'The bash command to execute. Can be empty string to view additional logs when previous exit code is `-1`. Can be `C-c` (Ctrl+C) to interrupt the currently running process. Note: You can only execute one bash command at a time. If you need to run multiple commands sequentially, you can use `&&` or `;` to chain them together.',
                },
            },
            'required': ['command'],
        },
    },
}

str_replace_editor = {
    'type': 'function',
    'function': {
        'name': 'str_replace_editor',
        'description': """Custom editing tool for viewing, creating and editing files in plain-text format
* State is persistent across command calls and discussions with the user
* If `path` is a file, `view` displays the result of applying `cat -n`. If `path` is a directory, `view` lists non-hidden files and directories up to 2 levels deep
* The `create` command cannot be used if the specified `path` already exists as a file
* If a `command` generates a long output, it will be truncated and marked with `<response clipped>`
* The `undo_edit` command will revert the last edit made to the file at `path`

Notes for using the `str_replace` command:
* The `old_str` parameter should match EXACTLY one or more consecutive lines from the original file. Be mindful of whitespaces!
* If the `old_str` parameter is not unique in the file, the replacement will not be performed. Make sure to include enough context in `old_str` to make it unique
* The `new_str` parameter should contain the edited lines that should replace the `old_str`
""",
        'parameters': {
            'type': 'object',
            'properties': {
                'command': {
                    'description': 'The commands to run. Allowed options are: `view`, `create`, `str_replace`, `insert`, `undo_edit`.',
                    'enum': ['view', 'create', 'str_replace', 'insert', 'undo_edit'],
                    'type': 'string',
                },
                'path': {
                    'description': 'Absolute path to file or directory, e.g. `/workspace/file.py` or `/workspace`.',
                    'type': 'string',
                },
                'file_text': {
                    'description': 'Required parameter of `create` command, with the content of the file to be created.',
                    'type': 'string',
                },
                'old_str': {
                    'description': 'Required parameter of `str_replace` command containing the string in `path` to replace.',
                    'type': 'string',
                },
                'new_str': {
                    'description': 'Optional parameter of `str_replace` command containing the new string (if not given, no string will be added). Required parameter of `insert` command containing the string to insert.',
                    'type': 'string',
                },
                'insert_line': {
                    'description': 'Required parameter of `insert` command. The `new_str` will be inserted AFTER the line `insert_line` of `path`.',
                    'type': 'integer',
                },
                'view_range': {
                    'description': 'Optional parameter of `view` command when `path` points to a file. If none is given, the full file is shown. If provided, the file will be shown in the indicated line number range, e.g. [11, 12] will show lines 11 and 12. Indexing at 1 to start. Setting `[start_line, -1]` shows all lines from `start_line` to the end of the file.',
                    'items': {'type': 'integer'},
                    'type': 'array',
                },
            },
            'required': ['command', 'path'],
        },
    },
}

think = {
    'type': 'function',
    'function': {
        'name': 'think',
        'description': """Use the tool to think about something. It will not obtain new information or make any changes to the repository, but just log the thought. Use it when complex reasoning or brainstorming is needed.

Common use cases:
1. When exploring a repository and discovering the source of a bug, call this tool to brainstorm several unique ways of fixing the bug, and assess which change(s) are likely to be simplest and most effective.
2. After receiving test results, use this tool to brainstorm ways to fix failing tests.
3. When planning a complex refactoring, use this tool to outline different approaches and their tradeoffs.
4. When designing a new feature, use this tool to think through architecture decisions and implementation details.
5. When debugging a complex issue, use this tool to organize your thoughts and hypotheses.

The tool simply logs your thought process for better transparency and does not execute any code or make changes.
""",
        'parameters': {
            'type': 'object',
            'properties': {
                'content': {'type': 'string', 'description': 'The content of your thought.'},
            },
            'required': ['content'],
        },
    },
}

finish = {
    'type': 'function',
    'function': {
        'name': 'finish',
        'description': """Finish the interaction when the task is complete OR if the assistant cannot proceed further with the task.""",
        'parameters': {
            'type': 'object',
            'properties': {
                'message': {
                    'type': 'string',
                    'description': 'A comprehensive message describing task completion, results achieved, any state changes made, key insights discovered, and other notes.',
                },
            },
            'required': [],
        },
    },
}
\end{Verbatim}
}
When the agent calls \code{finish}, the git diff is fetched from the Docker environment, and the reward is calculated by applying the git diff to the another Docker environment and running the unit tests.

\subsection{Context Folding}
For context folding, we implement these tools:
{\footnotesize
\begin{Verbatim}[breaklines, breakanywhere]
branch = {
    'type': 'function',
    'function': {
        'name': 'branch',
        'description': """Create a sub-branch to execute a sub-task.""",
        'parameters': {
            'type': 'object',
            'properties': {
                'description': {
                    'description': 'A concise 3-5 word identifier for the sub-task.',
                    'type': 'string'
                },
                'prompt': {
                    'description': 'Clear, compact task prompt: state objectives and critical info to preserve in the response. Be brief and informative.',
                    'type': 'string'
                },
            },
            'required': ['description', 'prompt'],
        },
    },
}
return_tool = {
    'type': 'function',
    'function': {
        'name': 'return',
        'description': """Finish the interaction when the sub task is complete OR if the assistant cannot proceed further with the task.""",
        'parameters': {
            'type': 'object',
            'properties': {
                'message': {
                    'type': 'string',
                    'description': 'A comprehensive message describing sub task outcome.',
                },
            },
            'required': ['message'],
        },
    },
}
\end{Verbatim}
}

The \code{branch} tool returns a template message, while the \code{return} tool rolls back the context to the previous turn that invoked the \code{branch} tool and appends a template message that repeats the \code{message} field.